\pdfoutput=1

\documentclass[11pt]{article}

\usepackage{EACL2023}

\usepackage{times}
\usepackage{latexsym}

\usepackage[T1]{fontenc}

\usepackage[utf8]{inputenc}

\usepackage{microtype}

\usepackage{inconsolata}

\usepackage{graphicx}

\usepackage[inline]{enumitem}

\usepackage{adjustbox}

\usepackage{amsmath}
\usepackage{amssymb}

\title{How {BERT} Speaks Shakespearean English?\\ Evaluating Historical Bias in Contextual Language Models}

\author{Miriam Cuscito \\
  Università degli Studi di Cassino e del Lazio Meridionale \\
  \texttt{miriam.cuscito@unicas.it}
\AND
  Alfio Ferrara \and 
  Martin Ruskov \\
  Università degli Studi di Milano \\
\texttt{\{firstname\}.\{lastname\}@unimi.it} \\ 
}

\begin{document}
\maketitle
\begin{abstract}
In this paper, we explore the idea of analysing the historical bias of contextual language models based on BERT by measuring their adequacy with respect to Early Modern (EME) and Modern (ME) English. In our preliminary experiments, we perform fill-in-the-blank tests with 60 masked sentences (20 EME-specific, 20 ME-specific and 20 generic) and three different models (i.e., BERT Base, MacBERTh, English HLM)~\footnote{\tiny \url{https://huggingface.co/bert-base-uncased}, \newline \url{https://huggingface.co/emanjavacas/MacBERTh}, \newline \url{https://huggingface.co/dbmdz/bert-base-historic-english-cased}}.  We then rate the model predictions according to a 5-point bipolar scale between the two language varieties and derive a weighted score to measure the adequacy of each model to EME and ME varieties of English.

\end{abstract}

\section{Introduction}
Contextual language models (CLMs) are deep neural language models which create contextualised word representations, in the sense that the representation for each word depends on the entire context in which it is used. That is to say, word representations are a function of the entire input sentence. Such models are usually pre-trained on large textual corpora and designed to have high predictive capabilities. This makes them closely tied to the domains on which they were trained and dependent on the infrastructure upon which they are based.

The presence of various biases in CLMs has been extensively studied, typically with the aim of proposing effective mitigation strategies~\cite{de-vassimon-manela-etal-2021-stereotype,ahn-oh-2021-mitigating,mozafari_hate_2020}. 
However, there are instances where the bias in certain CLMs is not necessarily negative. This is particularly true when the bias manifested in the language reflects its socio-temporal context. This bias could be advantageous for tasks that demand this socio-temporal staging ~\cite{corfield_fleeting_2022,diaz-faes_explicit_2023}.

In this paper, we explore the idea of analysing the bias by focusing on the major syntactic, semantic, and grammatical differences between two varieties of the English language: Early Modern (EME) and Modern (ME). More precisely, we propose a method and a measure of adequacy to test the adherence of CLMs to the natural language variety of interest. In particular, we assess the level of diachronic bias of three CLMs: Bert-Base-Uncased (BERT Base)~\cite{devlin-etal-2019-bert};
MacBERTh~\cite{manjavacas_adapting_2022}; and
Bert Base Historic English (English HLM).
In our preliminary experiments, we perform tests with 60 masked questions in which the models have the task to predict the masked word in the sentence. We then rate the proposed responses according to a 5-point bipolar scale between the two language variants and derive a weighted score from the response probabilities and their respective scores on the scale.

These results, although preliminary, might suggest a method applicable in the digital humanities when CLMs are employed for the analysis of historical corpora.

\section{Related Work}
\label{sec:background}

If it is true that language shapes culture while it is shaped by it~\cite{boroditsky_how_2011}, language models in general -- and CLMs in particular -- constitute a still partially covered mirror of this dual relationship. Not only can a CLM be tested based on its level of representativeness of the language to determine its reliability, but also it can tell us about linguistic, social, and historical phenomena that concern the culture tied to that specific language.
In other words, a CLM could be a valuable tool towards the expansion  of the broader social knowledge of a given culture, rightfully becoming part of the basic tools of Cultural Analytics discussed by Manovich~\cite{manovich_cultural_2020}. According to Bruner's~\cite{bruner_pragmatics_1984} pragmatic-cultural perspective, learning a language also means learning the cultural patterns associated with it. Similarly, analysing the language in its various realisations would mean having the opportunity to visualise the underlying cultural patterns.

Moreover, CLMs can be highly beneficial also for  philological~\cite{lit_among_2019}, pragmatic~\cite{ruskov_who_2023}, critical~\cite{diaz-faes_explicit_2023}, and literary work~\cite{piper_modeling_2023}. However, the effectiveness of these models depends on their ability to adapt to language specificity in its historical dimension. This is typically achieved by training models on historical text corpora. However, the difficulty of accessing large historical documentary collections means that the models available are still few and requires verifying whether they adapt effectively to the historical linguistic context.

BERT is a foundational contextual language model (CLM) which to date is the most widely adopted~\cite{montanelli_survey_2023}. A number of studies have explored different forms of bias in BERT~\cite{de-vassimon-manela-etal-2021-stereotype,ahn-oh-2021-mitigating,mozafari_hate_2020}.
Three BERT-based CLMs are of particular interest for our study: \begin{enumerate*}[label=\textit{(\roman*)}]
    \item Bert-Base-Uncased~\cite{devlin-etal-2019-bert}, created from a corpus of texts from Wikipedia and BookCorpus and a model of contemporary language, which we use as a control condition in our experiment;
    \item MacBERTh~\cite{manjavacas_adapting_2022}, pre-trained on texts from 1500 to 1950; and
    \item Bert-Base-Historic-English, pre-trained on contemporary texts and fine-tuned on historical texts from the 19th century to the present
\end{enumerate*}.



\section{Method}
\label{sec:method}

To evaluate the adequacy of CLMs on a test set, we define a temporal valence task consisting of a collection of test sentences, each with a masked token (i.e., word).
This is a typical fill-in-the-blank task, where the models are required to predict the masked token.
Formally, we consider the following three sets:
\begin{enumerate*}[label=\textit{(\roman*)}]
    \item we denote with $\mathcal{S}$ the set of all test \emph{sentences},
    \item with $\mathcal{V}$ we denote a set of \emph{vocabulary} words, and
    \item with $\mathcal{T} = \{-1, -0.5, 0, 0.5, 1\}  \subset \mathbb{R}$, we denote a 5-point bipolar \emph{temporal valence scale}, 
where $-1$ represents the farthest historical period and $1$ the closest to today.
\end{enumerate*}

With the above notation, for each of the masked sentences (denoted as $s \in \mathcal{S}$), we define a function $\rho: \mathcal{S} \rightarrow \mathcal{T}$ representing the \emph{sentence temporal valence score}. This function indicates the period from which the masked sentence is typical. 

Then, we calculate a \emph{token-in-sentence temporal valence score} $\sigma : \mathcal{V} \times \mathcal{S} \rightarrow \mathcal{T}$, indicating the score of a token substituting the sentence mask.

The mentioned \emph{temporal valence scores} are assigned arbitrarily according to the research hypotheses. Taking this study as an example, the criterion used to determine each score was the degree of alignment of certain sentences or tokens with a specific historical period on a philological-linguistic basis. Scholars wishing to delve into language study using this methodological approach can selectively choose the score to assign to their test set based on their specific research needs. The versatility of the proposed methodology is evident in its adaptability to a diverse array of fields of interest. This flexibility enables researchers to seamlessly integrate personalized metrics, ensuring a tailored approach to analysis without undermining the inherent consistency of the results.

As an example of \emph{temporal valence score}, given EME (Early Modern English) as the farthest period (i.e., $-1 \in \mathcal{T}$) and ME (Modern English) as closest (i.e., $1 \in \mathcal{T}$), if we consider the sentence $s_1 =$ \emph{``Why wilt [MASK] be offended by that?''} we have $\rho(s_1) = -1$ as $s_1$ is a representative sentence for EME, and $\sigma(``thou", s_1) = -1$, because in this context ``thou'' is indicative for EME. On the other hand, $\sigma(``not", s_1) = 0$, because ``not'' is neutral regarding the two language varieties.

Given a model $m$, for the masked token in each sentence ($s \in S$), we have the set of $\{w_1, w_2, \dots, w_n\} \subset \mathcal{V}$ of $n$ words predicted by $m$ for $s$, that are associated with the vector of corresponding probabilities \linebreak$\mathbf{p}_m = (p(w_1), p(w_2), \dots, p(w_n))^{T}$. 

For this set, we exploit the score function $\sigma$ in order to define a token-in-sentence temporal valence score vector $\mathbf{x}_m$ for $m$ given the sentence $s$, that is $\mathbf{x}_{m} = (\sigma(w_1, s), \sigma(w_2, s), \dots, \sigma(w_n, s))^{T}$.

This allows us to define the \emph{bias} of a model regarding the sentence as a weighted score:
$$\beta(m, s) = \mathbf{x}_{m}^{T}\mathbf{p}_m$$
Then, we proceed to define the \emph{domain adequacy} of a model with respect to a sentence $s$ as
$$\delta(m, s)=1-\frac{1}{2}\mid\rho(s)-\beta(m, s)\mid$$
based on the difference between the sentence temporal valence score $\rho(s)$ and the model bias $\beta(m, s)$. 

An example of three sentences from different periods is provided in Tables~\ref{tab:example-eme}, ~\ref{tab:example-neutral} and~\ref{tab:example-me}, which show the corresponding values for $\rho$, $p$, $\sigma$, $\beta(m,s)$ and $\delta(m,s)$.

\begin{table}[!h]
\centering
\begin{adjustbox}{width=\linewidth, keepaspectratio}  
\begin{tabular}{lrr}
\hline
\multicolumn{3}{c}{\textbf{BERT Base}}\\
\textbf{token} & \textbf{$p$} & \textbf{$\sigma$}\\
\hline
\verb|thou| & 0.712 & -1.0 \\ 
\verb|you| & 0.101 & 0.0 \\ 
\verb|i| & 0.085 & 0.0 \\ 
\verb|she| & 0.055 & 0.0 \\ 
\verb|he| & 0.048 & 0.0 \\ 
\hline
$\beta$ & \multicolumn{2}{c}{-0.712}\\
$\delta$ & \multicolumn{2}{c}{0.856}\\
\hline
\end{tabular}

\begin{tabular}{lrr}
\hline
\multicolumn{3}{c}{\textbf{MacBERTh}}\\
\textbf{token} & \textbf{$p$} & \textbf{$\sigma$}\\
\hline
\verb|thou| & 0.987 & -1.0\\
\verb|not| & 0.008 & 0.0 \\
\verb|you| & 0.004 & 0.0 \\ 
\verb|ye| & 0.001 & -1.0 \\ 
\verb|he| & 0.000 & 0.0 \\
\hline
$\beta$ & \multicolumn{2}{c}{-0.988}\\
$\delta$ & \multicolumn{2}{c}{0.994}\\
\hline
\end{tabular}

\begin{tabular}{lrr}
\hline
\multicolumn{3}{c}{\textbf{English HLM}}\\
\textbf{token} & \textbf{$p$} & \textbf{$\sigma$}\\
\hline
\verb|not| & 0.639 & 0.0 \\ 
\verb|thou| & 0.303 & -1.0 \\ 
\verb|never| & 0.031 & 0.0 \\ 
\verb|[UNK]| & 0.022 & 0.0 \\ 
\verb|ever| & 0.005 & 0.0 \\ 
\hline
$\beta$ & \multicolumn{2}{c}{-0.303}\\
$\delta$ & \multicolumn{2}{c}{0.652}\\
\hline
\end{tabular}
\end{adjustbox}
\caption{Scores of the models for ``Why willt [MASK] be offended by that?'' (temporal valence $\rho=-1$)}
\label{tab:example-eme}
\end{table}
\begin{table}[!h]
\centering
\begin{adjustbox}{width=\linewidth, keepaspectratio}  

\begin{tabular}{lrr}
\hline
\multicolumn{3}{c}{\textbf{BERT Base}}\\
\textbf{token} & \textbf{$p$} & \textbf{$\sigma$}\\
\hline
\verb|here| & 0.924 & 0.0 \\ 
\verb|back| & 0.066 & 0.5 \\ 
\verb|there| & 0.004 & -0.5 \\ 
\verb|forth| & 0.003 & -0.5 \\ 
\verb|out| & 0.003 & 1 \\ 
\hline
$\beta$ & \multicolumn{2}{c}{0.032}\\
$\delta$ & \multicolumn{2}{c}{0.984}\\
\hline
\end{tabular}

\begin{tabular}{lrr}
\hline
\multicolumn{3}{c}{\textbf{MacBERTh}}\\
\textbf{token} & \textbf{$p$} & \textbf{$\sigma$}\\
\hline
\verb|hither| & 0.740 & -1.0 \\
\verb|down| & 0.170 & 0.0 \\
\verb|thus| & 0.045 & -0.5 \\ 
\verb|in| & 0.025 & 0.0 \\ 
\verb|again| & 0.020 & 0.0 \\
\hline
$\beta$ & \multicolumn{2}{c}{-0.762}\\
$\delta$ & \multicolumn{2}{c}{0.619}\\
\hline
\end{tabular}

\begin{tabular}{lrr}
\hline
\multicolumn{3}{c}{\textbf{English HLM}}\\
\textbf{token} & \textbf{$p$} & \textbf{$\sigma$}\\
\hline
\verb|here| & 0.691 & 0.0 \\ 
\verb|back| & 0.194 & 0.5 \\ 
\verb|again| & 0.051 & 0.0 \\ 
\verb|in| & 0.034 & 0.0 \\ 
\verb|hither| & 0.031 & -1.0 \\ 
\hline
$\beta$ & \multicolumn{2}{c}{0.066}\\
$\delta$ & \multicolumn{2}{c}{0.967}\\
\hline
\end{tabular}
\end{adjustbox}
\caption{Scores for ``Have you come [MASK] to torment us before the time?'' ($\rho=0$)}
\label{tab:example-neutral}
\end{table}

\begin{table}[!h]
\centering
\begin{adjustbox}{width=\linewidth, keepaspectratio}  
\begin{tabular}{lrr}
\hline
\multicolumn{3}{c}{\textbf{BERT Base}}\\
\textbf{token} & \textbf{$p$} & \textbf{$\sigma$}\\
\hline
\verb|orientation| & 0.720 & 1.0 \\ 
\verb|misconduct| & 0.112 & 1.0 \\ 
\verb|minorities| & 0.067 & 1.0 \\ 
\verb|partners| & 0.052 & 1.0 \\ 
\verb|harassment| & 0.048 & 1.0 \\ 
\hline
$\beta$ & \multicolumn{2}{c}{1.000}\\
$\delta$ & \multicolumn{2}{c}{1.000}\\
\hline
\end{tabular}

\begin{tabular}{lrr}
\hline
\multicolumn{3}{c}{\textbf{MacBERTh}}\\
\textbf{token} & \textbf{$p$} & \textbf{$\sigma$}\\
\hline
\verb|##ists| & 0.493 & -0.5 \\
\verb|offenders| & 0.165 & 1.0 \\
\verb|characters| & 0.130 & 0.5 \\ 
\verb|drunkards| & 0.117 & 0.0 \\ 
\verb|delinquents| & 0.095 & 0.5 \\
\hline
$\beta$ & \multicolumn{2}{c}{0.031}\\
$\delta$ & \multicolumn{2}{c}{0.516}\\
\hline
\end{tabular}

\begin{tabular}{lrr}
\hline
\multicolumn{3}{c}{\textbf{English HLM}}\\
\textbf{token} & \textbf{$p$} & \textbf{$\sigma$}\\
\hline
\verb|to| & 0.319 & 0.0 \\ 
\verb|must| & 0.294 & 0.0 \\ 
\verb|may| & 0.187 & 0.0 \\ 
\verb|would| & 0.104 & 0.0 \\ 
\verb|should| & 0.096 & 0.0 \\ 
\hline
$\beta$ & \multicolumn{2}{c}{0.000}\\
$\delta$ & \multicolumn{2}{c}{0.500}\\
\hline
\end{tabular}
\end{adjustbox}
\caption{Scores for ``Should men who are known sexual [MASK] be given a platform?'' ($\rho=1$)}
\label{tab:example-me}
\end{table}





\section{Evaluation}
\label{sec:results}

We test our metrics with three BERT-based linguistic models we consider relevant for the varieties of the English language of interest: \emph{(i)} Bert-Base-Uncased, \emph{(ii)} MacBERTh, and \emph{(iii)} Bert-Base-Historic-English. In accordance with the objectives of this study, the choice of models reflects a specific interest in language; therefore, they can be replaced to best fit any other specific interest in diachronic language analysis. 
For the test we used 60 word-masked sentences, specifically created for this study. To create the test set, we relied on different types of written language: contemporary standard, journalistic language, social media non-standard, and Early Modern language.

The elements to be masked were selected based on their belonging to specific word classes known to have suffered more exposure to the diachronic variation of the English language: pronouns, verbs, adverbs, adjectives, and nouns. Of the 60 sentences provided in Appendix, 20 are selected to be suggestive for the EME variety of English, further 20 – as suggestive for ME, and final 20 are generic. Once the test set was complete, a \textit{temporal valence score} was assigned to each sentence (see $\rho$ in Section~\ref{sec:method}) based on their level of chronological markedness.

The test set was administered to the three CLMs, and the suggested words with their probability were collected. The resultant vocabulary was marked independently from the models that provided it by setting the \textit{token-in-sentence temporal valence score} (i.e. $\sigma$) to each word, based on an estimation of proximity of the token's meaning to a certain linguistic variety in the context in which it appeared. Notably, during this phase, our decision was to work on a sentence level (contextually) rather than on a set level (globally). The method proved highly effective in avoiding the risk of semantic flattening, given that almost every word has shown some level of contextual semantic specificity if taken contextually rather than globally. An example is the pronoun \textit{you} in “\textit{fare you well, sir”}, which is globally neutral and yet acquires a strong diachronic value if evaluated in its context, in which it appears to be utmost archaic. 

\begin{figure*}[!h]
  \centering
  \includegraphics[width=\textwidth]{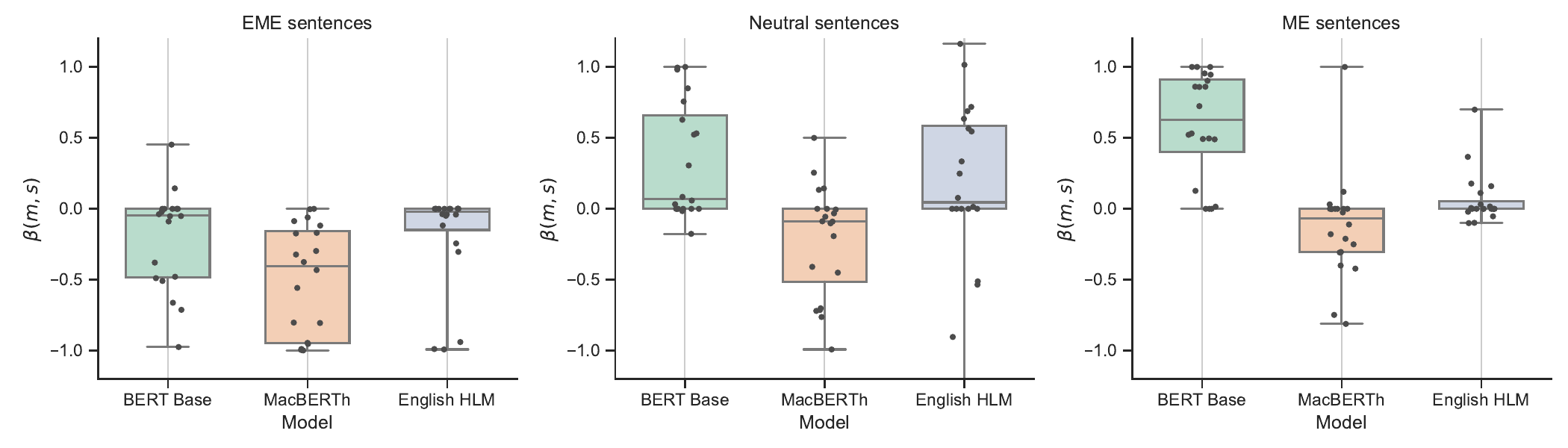}
  \caption{Distribution of bias $\beta(m, s)$ of the three models with respect to the three test sets.}
  \label{fig:beta}
\end{figure*}

\begin{figure*}[!h]
  \centering
  \includegraphics[width=\textwidth]{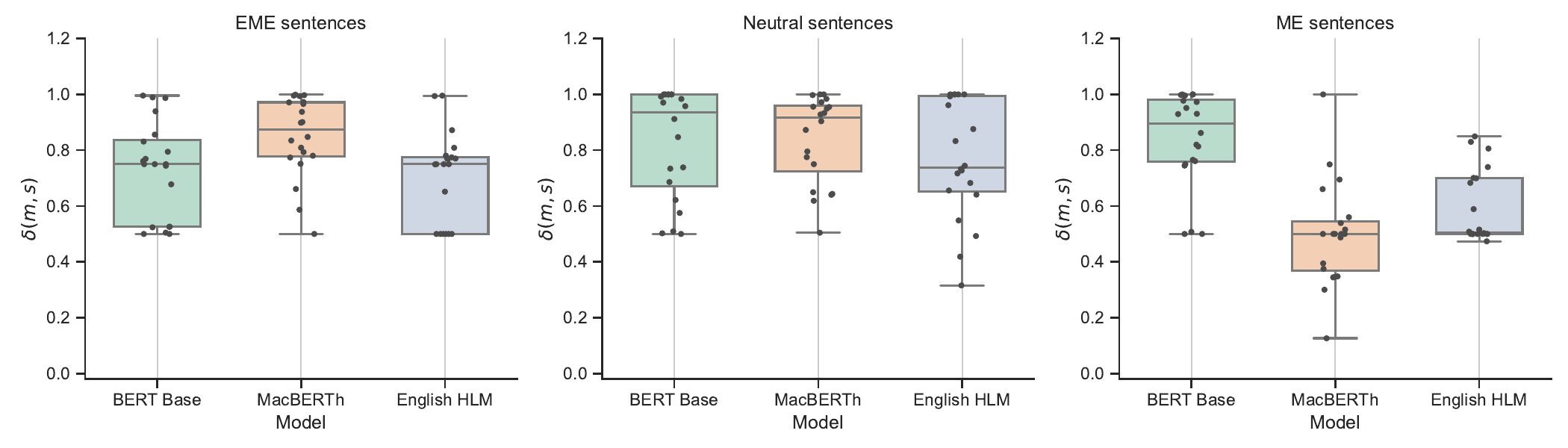}
  \caption{Distribution of the domain adequacy $\delta(m, s)$ of the three models with respect to the three test sets.}
  \label{fig:delta}
\end{figure*}

Once $\beta$ and  $\delta$ were calculated, we proceeded with the analysis of the data and the collection of results. The distribution of the bias score $\beta$ and the domain adequacy score $\delta$ for the sentences in the three groups (i.e., EME, Neutral, and ME) is shown in Figures~\ref{fig:beta} and~\ref{fig:delta}, respectively.

Figure~\ref{fig:beta} shows that for all three test sets, MacBERTh is most aligned with EME, whereas BERT Base is always most aligned as ME. Historical BERT shows a tendency towards a more neutral language than the other two models in marked sentences, whilst surprisingly it aligns to ME in neutral sentences.

Figure~\ref{fig:delta} shows that MacBERTh has best domain adequacy for EME, and BERT Base has best domain adequacy for ME.
In the case of the neutral test set, domain adequacy is no less informative. Although the sentences do not inherently carry their expectations regarding language, models appear consistently well-suited to a neutral context, and none of them pushes for strong specificity of their corresponding trained domain. In effect this leaves the sentences close to their original neutrality

This preliminary study provides an illustration of the nature and functioning of the CLMs predictive behaviour. The presence or absence of markedness in the sentences enables all three CLMs to select the type of element which best fits the co-text. So, while for diachronically marked sentences, models without training in that domain attempted to suggest probable solutions, sometimes resulting in a form of linguistically inconsistent mimicry, in unmarked sentences, the models perform exceptionally well, and linguistic inaccuracies are rare.

\section{Conclusion}
\label{sec:conclusion}

Both notions of \emph{bias} ($\beta$) and \emph{domain adequacy} ($\delta$) provide important insights of the nature of the models. The first, $\beta$, indicates a tendency in terms of temporal valency. In other words, the interpretation of its value should be considered within the context of the specific dichotomy of language varieties. On the other hand, $\delta$ reflects the adequacy for an individual language variety. It successfully captures model tendencies when completing historically predetermined sentences. However, it falls short in capturing the preferred language variety when completing temporality-neutral sentences.

The notion that LMs can serve as a window into the history of a population is not new, but there is a growing interest in exploring the relationships between these models and the sociolinguistic and sociocultural contexts. It is equally imperative to establish a procedural framework to address the lack of evaluative methods for these models, as previously hinted at in this text.

Within this evaluation, we created a dedicated test set for each model under scrutiny, drawing upon approaches used for evaluation of bias in CLMs. In creating our test sets, we built our sentences both on logical-semantic and logical-syntactic tasks. Future work could try to create a test set for model interrogation that is culture-oriented, delving into socio-culturally significant elements such as customs, historical events, and attitudes towards social groups – elements recognised as belonging to social knowledge.

This study aims not only to propose a methodology for assessing language models but also to put forth hypotheses for expanding the available tools  to humanities scholars interested in studying complex socio-cultural phenomena with an approach which begins by interpreting textual clues and inferring their connections to reality.

\bibliography{anthology,custom,oed}
\bibliographystyle{acl_natbib}

\appendix

\section{Test Set}
\label{sec:appendix}

For transparency and reproducibility purposes, the following anonymous link contains the complete test set with the corresponding values produced during evaluation:\\
\url{https://tinyurl.com/bert-shakespearean}

\end{document}